# A Computational Algorithm based on Empirical Analysis, that Composes Sanskrit Poetry

Rama N.
Department of Computer Science
Presidency College
Chennai, India
n_ramabalu@yahoo.com

Meenakshi Lakshmanan
Department of Computer Science
Meenakshi College for Women
Chennai, India
and
Research Scholar, Mother Teresa Women's University
Kodaikanal, India

*Abstract* — **Poetry-writing in Sanskrit is riddled with problems for even those who know the language well. This is so because the rules that govern Sanskrit prosody are numerous and stringent.**

**We propose a computational algorithm that converts prose given as E-text into poetry in accordance with the metrical rules of Sanskrit prosody, simultaneously taking care to ensure that *sandhi* or euphonic conjunction, which is compulsory in verse, is handled. The algorithm is considerably speeded up by a novel method of reducing the target search database. The algorithm further gives suggestions to the poet in case what he/she has given as the input prose is impossible to fit into any allowed metrical format. There is also an interactive component of the algorithm by which the algorithm interacts with the poet to resolve ambiguities. In addition, this unique work, which provides a solution to a problem that has never been addressed before, provides a simple yet effective speech recognition interface that would help the visually impaired dictate words in E-text, which is in turn versified by our Poetry Composer Engine.**

I. INTRODUCTION

Poetry-writing in any language has always posed a challenge, causing poets to be acclaimed as a lofty tribe. The case is particularly strengthened when it comes to writing Sanskrit poetry, which is subject to numerous stringent rules at the grammatical, semantic and metrical levels, with compound word formations and euphonic conjunctions (*sandhi*-s) exacerbating the complexity.

In Sanskrit poetry, words may be positioned anywhere and in any order in the verse, and need not read in prose-order. The prose-order is later gleaned by bunching together words of the same genre, i.e. those belonging to the same case-inflectional form, gender, number or tense. For example, an adjective can easily be matched with the noun it describes, by looking for the closest noun that is in the same case-inflectional form, gender and number as itself. This re-organizing into the prose-order is the procedure followed in practice in order to decipher the meaning of a verse.

The computational algorithm we propose, converts prose given as E-text into poetry in strict accordance with the metrical rules of Sanskrit prosody. What would otherwise be an algorithm with very high time complexity, is considerably improved by a novel scheme by which the target search space is narrowed down significantly. The algorithm also handles *sandhi*-s, since they are compulsorily to be implemented while versifying. Another path-breaking feature of the algorithm is that it gives suggestions to the poet in case his/her input cannot be versified according to the metrical rules. Moreover, in cases where the *sandhi* rule is ambiguous, the algorithm interacts with the poet to confirm his decision on resolving the same. In addition, a speech recognition interface is also provided, that would help the visually impaired dictate words in E-text, which are then versified by the Poetry Composer Engine.

*A. Unicode Representation of Sanskrit Text*

The Unicode (UTF-8) standard is what has been adopted universally for the purpose of encoding Indian language texts into digital format. The Unicode Consortium has assigned the Unicode hexadecimal range 0900 - 097F for Sanskrit characters.

All characters including the diacritical characters used to represent Sanskrit letters in E-texts are found dispersed across the Basic Latin (0000-007F), Latin-1 Supplement (0080-00FF), Latin Extended-A (0100-017F) and Latin Extended Additional (1E00 – 1EFF) Unicode ranges.

The Latin character set has been employed in this paper to represent Sanskrit letters as E-text.

The text given in the form of E-text using the Unicode Latin character set, is taken as input for processing. Unicode Sanskrit font may also be accepted as input, but is converted to the Latin character form before processing begins, as already presented by the authors in [3].

II. RULES OF VERSIFICATION

*A. Metrical Rules*

Verses in Sanskrit are classified according to metres, i.e. according to the number and type of syllables in the four quarters of the verse (and in a few cases, in the two halves of the verse). Algorithms to efficiently parse and classify verses into more than 700 metres, have already been developed by the authors [3]. The following is a brief on the overall classification scheme and methodology.

Sanskrit verse is normally taken to consist of a sequence of four *pāda*-s or quarters. Each quarter is regulated either by the number of syllables (*akṣara*-s) or the number of syllabic instants (*mātrā*-s) and the determination of metres is based on either of these factors [3]. Metres based on the first yardstick are called *varṇa* metres, while those based on the second are termed *jāti* metres.

*Varṇa* **Metres**

A syllable is as much of a word as can be pronounced at once. There are two types of syllables to contend with – the long (*guru*) and the short (*laghu*). The following is how syllables are categorized as long and short:

Short syllables:

- Normally, all short vowels – *a, i, u, ṛ, ḷ.*

Long syllables:

- All long vowels – *ā, ī, ū, ṝ.*
- Any short vowel followed by the *anusvāra* (*ṁ*).
- Any short vowel followed by the *visarga* (*ḥ*).
- Any short vowel followed by a double consonant. (The exceptions to this rule are the double consonants *pr, br, kr* and those starting with *h*. In these four cases, the preceding short vowel can optionally remain short.)
- Optionally, any short vowel at the end of a *pāda*.

The optional nature of the exceptions mentioned in the last two rules above, indicates a sort of poetic license.

From the above discussion it is clear that the four quarters of a verse can each be represented as a sequence of long and short syllables. Traditionally, identification of *varṇa* metres is done on the basis of metrical feet, termed '*gaṇa*-s' in Sanskrit. A *gaṇa* is a combination of three syllables, each of which may be long or short. As such, there are eight such *gaṇa*-s defined as in Table I, in which '*l*' stands for a *laghu* (short) letter, and '*g*' for a *guru* (long) one.

The number of syllables in a quarter can vary from 1 to 999. When the number of syllables is between 1 and 26 per quarter, the meters are categorized into three:

a. *Sama* (meaning 'equal') – In this, all the four quarters of the verse are identical not just in terms of the number of syllables, but also in the sequence of long and short syllables

b. *Ardhasama* (meaning 'half-equal') – In this, the first and third quarters are identical, as are the second and fourth.

c. *Viṣama* (meaning 'unequal') – In this, the quarters are uneven or mixed up.

TABLE I. *GAṆA* SCHEME

| # | Syllable-combination | *Gaṇa* | Corresponding English Category |
|---|---|---|---|
| 1 | *lgg* | *y* | Bacchius |
| 2 | *glg* | *r* | Amphimacer |
| 3 | *ggl* | *t* | Anti-bacchius |
| 4 | *gll* | *bh* (denoted as *b*) | Dactylus |
| 5 | *lgl* | *j* | Amphibrachys |
| 6 | *llg* | *s* | Anapaestus |
| 7 | *ggg* | *m* | Molussus |
| 8 | *lll* | *n* | Tribrachys |

The meters in which there are greater than 26 syllables per quarter are of the '*daṇḍaka*' type and are beyond the scope of this work. [3]

Given that each syllable can be either '*l*' or '*g*', there is clearly a combinatorial explosion in the number of possible 1-syllabled to 26-syllabled *Sama* metres. For *Ardhasama* metres the possible number is obviously even higher, and with *Viṣama*, the possibilities are infinite. However, the number of metres in actual use across the literature is limited to a smaller number than the number theoretically possible. Hence, it is sufficient to handle the metres in vogue [4]. An example is given in Table II.

TABLE II. L-G AND *GAṆA* REPRESENTATIONS OF A SAMPLE VERSE

| Verse | *l-g* syllables | *Gaṇa* |
|---|---|---|
| *vande gurūṇāṁ caraṇāravinde* | ggl ggl lgl gg | ttjgg |
| *sandarśitasvātmasukhāvabodhe \|* | ggl ggl lgl gg | ttjgg |
| *janasya ye jāṅgalikāyamāne* | lgl ggl lgl gg | jtjgg |
| *saṁsārahālāhalamohaśāntyai \|\|* | ggl ggl lgl gg | ttjgg |

The metre of a verse having the *gaṇa* sequence "*ttjgg*" in all its four *pāda*-s is called "*Indravajrā*", while that of a verse having the *gaṇa* sequence "*jtjgg*" in all its four *pāda*-s is called "*Upendravajrā*". A verse such as the example given in Table II, whose *gaṇa* sequences are a combination of the *Indravajrā* and *Upendravajrā* sequences, is said to have the metre called "*Upajāti*".

*Jāti* **Metres**

In this type of metre, each short syllable is counted as constituting one syllabic foot or *mātrā*, while a long syllable is counted as constituting two. Such metres are categorized into two, depending on whether the verse is considered as constituted by two halves, or by four quarters. The various types of the oft-encountered *Āryā* metres are examples of the first variety.

The identification of metres is done based mainly on the number of certain groups of *mātrā*-s and also partially on *gaṇa* patterns. Standard groups of *mātrā*-s are those of 4, 6, 8 and 10 *mātrā*-s. [3]

While composing verse, a poet must make sure that the length of a pāda or half-verse (as the case may be), falls into one of the accepted metre types, either *varṇa* or *jāti*. This implies that the *laghu-guru* (L-G) combinations of syllables in the verse comply with some accepted metre-format. This forms the main challenge in composing poetry in Sanskrit, and thus has to be most importantly ensured while creating poetry out of a given Sanskrit sentence or group of sentences.

*B. Rules of Euphonic Conjunctions*

Euphonic conjunctions or *sandhi*-s in Sanskrit are points between adjacent words or sub-words at which letters coalesce and transform. The application of *sandhi* is compulsory in Sanskrit verse, though the rules are not as stringent in the prose. A novel computational approach to *sandhi* processing based on building *sandhi*-s rather than splitting them, was developed by the authors [2]. This was done in accordance with the grammatical rules laid down by the ancient Sanskrit grammarian-genius Pāṇini in his magnum opus, the *Aṣṭādhyāyī* and forms a comprehensive *sandhi*-building engine.

An example of *sandhi* is: *tat* + *ṭīkā* = *taṭṭīkā*. This is an example of the *ṣṭutva sandhi* type of euphonic conjunctions [2]. In this case, the word *tat* is an '*l*' syllable, while the word *ṭīkā* maps to '*gg*'. Hence, with these two words given in the above sequence, we get the *l-g* sequence, "*lgg*". Now application of the *sandhi* rule as shown above causes a transformation of the word. As per the rules outlined in section 3.1 above, the word *taṭṭīkā* has the *l-g* sequence "*ggg*". Thus, the first syllable provided by the word *tat* gets transformed from short type to long type.

While composing poetry by changing the order of the given words, care must be taken to handle all such *sandhi* rules that affect the long and short syllables.

There is, however, an exception to *sandhi* rules that must be handled as well. Normally, *sandhi* rules operational on *phale* + *atra* would transform into *phale'tra*, i.e. the starting vowel '*a*' of the second word would be dropped. Now the word *phale* can have two meanings: 'in the fruit' and 'two fruits'. In the latter case, i.e. when the word is in the dual number, the above *sandhi* rule will not be operational.

### III. THE PROSE-TO-POETRY CONVERTER ALGORITHM

*A. The Input*

The input is given as E-text with diacritical marks, using the Latin Unicode character set mentioned in Section 1.1. It is assumed that compound words are properly formed before the input is given. For example, the compound word *gaṅgājalam* is actually composed of two words *gaṅgāyāḥ jalam* meaning "(River) Gaṅgā's water". So, if the poet giving the input wishes to indicate this meaning, he would have to give '*gaṅgāyāḥ*' and '*jalam*' as two separate words, or he would have to provide the compound word '*gaṅgājalam*'. Giving the input as the two words '*gaṅgā*' and '*jalam*' would be incorrect and unacceptable, unless the poet wants to refer to River Gaṅgā independently and water independently, without intending to convey the relationship "Gaṅgā's water".

To facilitate ease of access for the visually impaired, a simple alternative user interface is introduced. Since English style E-text is what is employed for input, the application is made voice-enabled, so that the input can be called out letter by letter, providing a pause between words. For example, to input the word "*bhagavān*", the user calls out, 'b', 'h', 'a', 'g', 'a', 'v', 'A', 'n' [6]. The special diacritical characters like '*ā*' are processed specially by the input module, as shown in Table III. Converting the output verse form to speech for the benefit of such users, has already been solved by the authors in earlier work [4].

TABLE III. SCHEME OF CALLING OUT SPECIAL CHARACTERS FOR VOICE INPUT

| # | Diacritical letter | English speech equivalent (all in capital) |
|---|---|---|
| 1 | *ā* | A |
| 2 | *ī* | I |
| 3 | *ū* | U |
| 4 | *ṛ* | R |
| 5 | *ṝ* | F |
| 6 | *ḷ* | L |
| 7 | *ṅ* | G |
| 8 | *ñ* | Y |
| 9 | *ṭ* | T |
| 10 | *ḍ* | D |
| 11 | *ṇ* | N |
| 12 | *ś* | S |
| 13 | *ṣ* | Z |
| 14 | *ḥ* | H |
| 15 | *ṁ* | M |

*B. The Overall Algorithmic Approach*

The following is the overall algorithm for converting the given input into poetry.

**Algorithm** ComposePoetry

**Step 1:** Scan the given text and establish the maximum and minimum possible number of syllables in a verse that can result from the given input. This is done as follows:

**Step 1.1: Determination of Maximum (Max): Max =** the total number of vowels in the input.

Note: The number n denotes the maximum number of syllables possible in the verse, no matter in what order the words are jumbled up, and no matter what *sandhi* rules have to consequently be applied. This is because the application of *sandhi* rules can only reduce the number of syllables and can never increase them.

**Step 1.2: Determination of Minimum (Min):** Calculate **Min**, the minimum number of syllables in the verse, by examining possible *sandhi*-s that are applicable with the words in the given input, and the maximum number of reductions possible through them.

**Step 2:** Use the **Max-Min** band to reduce the set of possible metres that are likely for the given input.

**Step 3:** Convert each word in the input into its *l-g* equivalent.

**Step 4:** Starting from the original permutation of words,

for all possible permutations of the words, do apply sandhi rules and reduce;

if number of syllables after reduction is divisible by 4, then split the verse into equal-sized quarters;

```
        if the l-g patterns of the four quarters are equal then
            search for a match for the pattern in the reduced
            table of Sama metres;
        else if the alternate quarters have same l-g pattern then
            search for a match for the pattern in the reduced
            table of Ardhasama metres;
        else
            search for a match for the pattern in the reduced
            tables of Viṣama and Jāti metres;
        end if
        if match found then
            quit;
        else
            indicate the closest match;
            suggest possible changes;
        end if
    else
        split the verse according to possible Ardhasama metres;
        search for a match for the pattern in the reduced table of
        Ardhasama metres;
        if match found then
            quit;
        else
            split the verse according to possible Viṣama metres
            and Jāti metres;
            search for a match for the pattern in the reduced
            tables of Viṣama and Jāti metres;
            if match found then
                quit;
            else
                indicate the closest match;
                suggest possible changes;
            end if
        end if
    end if
end for
end Algorithm
```

The following points are worthy of note with regard to the above algorithm:

a. Giving big words as input is advantageous and would yield better performance, because the number of possible permutations would reduce.

b. The given order of words in the prose text is first tried as such, because the order of words given would be meaningful, and it would be ideal if a verse form is possible in the given order itself. Other permutations of words are tried only if a verse form is not possible with the given word-order.

c. The algorithm suggests possible changes by using other words available in the input with the required pattern.

d. In case the number of syllables is greater that 26 per quarter, then the given words are split into more than one set and then *Anuṣṭhup* (common metres with 8 syllables per quarter) or *Jāti* metres are tried out for a match.

IV. ESTABLISHING THE MAX-MIN BAND

The algorithm presented in Section 4.1 involves permutation-generation and string-matching for each permutation [1]. This fact coupled with the big size of the metres database, places rather large demands on time. It is to significantly enhance the performance of the algorithm, that the novel idea of establishing the maximum-minimum band has been devised.

Once this **Max-Min** band is established, metres with the number of syllables lying within this band alone need to be considered for pattern matching. This approach clearly ensures a substantial savings in terms of time taken by the algorithm.

TABLE IV. THE SANSKRIT ALPHABET CATEGORIZED

| # | Category | Letters |
|---|---|---|
| 1 | Vowels | *a, ā, i, ī, u, ū, ṛ, ṝ, ḷ, e, ai, o, au* |
| 2 | Short vowels | *a, i, u, ṛ, ḷ* |
| 3 | Long vowels | *ā, ī, ū, ṝ, e, ai, o, au* |
| 4 | Consonants (including Semi-vowels) | k, kh, g, gh, ṅ<br>c, ch, j, jh, ñ<br>ṭ, ṭh, ḍ, ḍh, ṇ<br>t, th, d, dh, n<br>p, ph, b, bh, m<br>y, r, l, v |
| 5 | Sibilants | ś, ṣ, s |
| 6 | Aspirate | h |
| 7 | *Anusvāra* | ṁ |
| 8 | *Visarga* | ḥ |

The algorithm presented in Section 4.1 defines how to find the maximum number of syllables possible in a verse formed from the given input. The following discussion focuses on the development and establishment of equations through comprehensive empirical analysis, to determine the minimum possible number of syllables in a verse constructed from the given input text. Table IV presents the letters of the Sanskrit alphabet divided into categories referenced in the discussion below.

Consider each individual word of the given input text. The observation we first make is that we have to contend with only those *sandhi*-s which can reduce vowels from long to short, for this is what will affect metres. Now we introduce the following possibilities and categorizations:

1. **Set $S_1$**: Words that begin with a vowel and end with a consonant, sibilant, aspirate, *anusvāra* or *visarga* (Eg: *ahaṁ*).

   The maximum change that such a word can bring about is 1, because it has only one vowel at its beginning which may merge with another word and get reduced.

   Let the cardinality of $S_1$ be $n_1$.

2. **Set $S_2$**: Words that begin with a consonant, sibilant or aspirate and end with a vowel other than *au, ai* (Eg: *bhavāmi*).

   The maximum change that such a word can bring about is 1, because it has only one vowel at its end which may merge with another word and get reduced.

   Let the cardinality of $S_2$ be $n_2$.

   The reason for not considering words ending with the long vowels *ai* and *au*, is that these can never, under the effect of any *sandhi* rule whatsoever, lead to a reduction from long to short. For example, in *kau + ūrudvayaṁ = kāvūrudvayaṁ*. the long vowel *au* does undergo a replacement, but only by another long vowel, viz. *ā*. As such, there is really no change in the *l-g* scheme that can be brought about by words in this category ending with *ai* and *au*. Hence we leave them out, thereby reducing the processing time further.

3. **Set $S_3$**: Words that begin and end with a vowel (Eg: *atra*).

   The maximum change that such a word can bring about is 2, because the vowels at both it ends may merge with adjacent words and get reduced.

   Let the cardinality of $S_3$ be $n_3$.

4. **Set $S_4$**: Words that begin with a consonant, sibilant or aspirate and end with a consonant, sibilant, aspirate, *anusvāra* or *visarga* (Eg: *marut*)

   Such words can effect no change at all, because no vowel in them gets reduced. Neither can they reduce any vowel in adjacent words. Hence we do not introduce any notation for the size of this set.

5. **Set $S_5$**: Words that begin with a vowel and end with "*aḥ*" (Eg: *ambaraḥ*).

   Clearly, this is a subset of $S_1$, and can cause a maximum change of 1.

   Let the cardinality of $S_5$ be $n_5$.

6. **Set $S_6$**: The word *ahaḥ* is special when it combines with another instance of itself, because *ahaḥ + ahaḥ = aharahaḥ*, which causes a change of 1.

   Clearly, this is a subset of $S_5$, and its cardinality is included in $n_5$.

7. **Set $S_7$**: Words that begin with a consonant, sibilant or aspirate and end with "*aḥ*" (Eg: *kṛṣṇaḥ*)

   Let the cardinality of $S_7$ be $n_7$.

We now derive partial equations for the maximum number of reductions possible in various scenarios, based on the above notations for number of word-occurrences. This maximum number is denoted by $r_m$ where m = 1, 2, 3.

*A. Formula for $r_1$ (incorporation of $S_1$, $S_2$ and $S_3$ words)*

   if $n_1 = 0$ and $n_2 = 0$ and $n_3 > 0$ then

   $r_1 = n_3 - 1$;

   else

   $r_1 = \min(n_1, n_2) + n_3$;

   end if

**Explanation**

In case both $S_1$ and $S_2$ are null sets, $S_3$ words can combine with only $S_3$ words, provided $S_3$ is not null. Hence, clearly, the maximum number of reductions possible is only $n_3 - 1$.

Consider the case when both $S_1$ and $S_2$ are not null sets. For words of the $S_1$ and $S_2$ categories, the maximum change can be brought about by aligning a $S_1$ word just after a $S_2$ word, whereby the vowels will combine and may reduce. That is, we pair them up. Hence, if $n_1$ and $n_2$ are unequal, then $\text{abs}(n_1-n_2)$ words will remain unpaired and will therefore cause no change. Hence, if $n_1$ and $n_2$ are non-zero, the maximum number of reductions possible is $\min(n_1, n_2)$.

As for words of the $S_3$ category, both ends may cause a change. Hence, irrespective of $n_1$ and $n_2$, $n_3$ number of changes will take place, provided $n_3$ is non-zero. Hence we add $n_3$ to the formula. For example, consider the following sentence provided as input:

*idānīṁ atra ālasyaṁ tyaktvā ahaṁ paṭhāmi ca likhāmi ca*

Here,

$S_1$ words: *idānīṁ, ālasyaṁ, ahaṁ*

$S_2$ words: *tyaktvā, paṭhāmi, ca, likhāmi, ca*

$S_3$ words: *atra*

Thus, $n_1 = 3$, $n_2 = 5$, $n_3 = 1$. Clearly there are only a maximum of 3 ways, i.e. $\min(n_1, n_2)$ ways, of pairing the $S_1$ and $S_2$ words to cause a change. Further, though both ends of the $S_3$ words can cause change, they can cause only one change each by combining with any other $S_3$ word or with any of the remaining words of $S_1$ or $S_2$. Hence we add $n_3$.

In the case where exactly one of $S_1$ and $S_2$ is a null set, then too, this formula will clearly hold, since $\min(n_1, n_2) = 0$ and the maximum possible number of reductions will hence simply be $n_3$, irrespective of the value of $\text{abs}(n_1-n_2)$.

*B. Formula for $r_2$ (incorporation of $S_5$ words)*

   if $n_1 = n_2$ and $n_5 > 0$ then

   $r_2 = n_5 - 1$;

   else

$r_2 = n_5$;
end if

**Explanation**

We consider the two cases when $n_1$ and $n_2$ are equal, and when they are unequal. When they are equal, they pair up completely for $r_1$, and hence if $S_5$ is not null, the $S_5$ words have only themselves to combine with. For example, we may have the $S_5$ words, *itaḥ* and *aṁbaraḥ*. Here, *itaḥ + aṁbaraḥ = ito'mbaraḥ*, and *aṁbaraḥ + itaḥ = aṁbara itaḥ*, both of which are changes. However, the first does not cause any change in the long-short scheme, while the second does. Since we are only calculating the maximum, we take it that there is a maximum reduction of 1 in this case. Clearly, therefore, the maximum number of reductions here is $n_5 - 1$.

In the case where $n_1$ and $n_2$ are unequal, they pair up to reduce as per $r_1$, leaving behind abs($n_1-n_2$) words. Consider the example,

*aṁbaraḥ na atra asti parantu aṁbaraḥ anyatra asti ataḥ ahaṁ itaḥ tatra gacchāmi*

Here,

$S_1$ words: *ahaṁ*

$S_2$ words: *na, parantu, tatra, gacchāmi*

$S_3$ words: *atra, asti, anyatra, asti*

$S_5$ words: *aṁbaraḥ, aṁbaraḥ, ataḥ, itaḥ*

Thus, $n_1 = 1$, $n_2 = 4$, $n_3 = 4$, $n_5 = 4$. Assuming that the first words of $S_1$ and $S_2$ combine, we have the last three words of $S_2$ left behind. These can produce three pairs in combination with three words of $S_5$ and cause a change. For example, we can have *tatra + aṁbaraḥ = tatrāmbaraḥ*. We would thus have one word of $S_5$ left behind, since $n_5 >$ abs($n_1-n_2$), which would combine with one of the available compounds of $S_5$ with $S_1$ or $S_2$. Continuing with the above example, *tatrāmbaraḥ + itaḥ = tatrāmbara itaḥ*, a change. All this means that all the $S_5$ words contribute to a change, and hence $r_2 = n_5$ in this case.

In case $n_1$ and $n_2$ are unequal and $n_5 <=$ abs($n_1-n_2$), then after pairing $S_5$ words with the remaining from $S_1$ or $S_2$, we are left with no more $S_5$ words. Hence, anyway the number of reductions is $n_5$, i.e. the number of pairs. The only other case is $n_5 = 0$, in which $r_2$ should work out to zero. This possibility is subsumed in the second case presented in the formula.

*C. Formula for $r_3$ (incorporation of $S_7$ words)*

if $n_1 > n_2$ and $n_7 > 0$ then

$r_3 = \min(n_7, n_1 - n_2)$;

else

$r_3 = 0$;

end if

**Explanation**

Clearly, we have to handle only the case when $n_7 > 0$. Also, after calculation of $r_1$, the remaining abs($n_1 - n_2$) words alone have to be contended with in the calculation of $r_3$.

Now $S_2$ words are of no use for reduction in combination with $S_7$ words. Consider the following sample input:

*kṛṣṇaḥ idānīṁ atra ālasyaṁ tyaktvā paṭhati ca likhati ca*

Here,

$S_1$ words: *idānīṁ, ālasyaṁ*

$S_2$ words: *tyaktvā, paṭhati, ca, likhati, ca*

$S_3$ words: *atra*

$S_7$ words: *kṛṣṇaḥ*

Thus, $n_1 = 2$, $n_2 = 5$, $n_3 = 1$, $n_7 = 1$. Clearly, after the $S_1$ and $S_2$ words combine, three $S_2$ words would remain, say *ca, likhati* and *ca*. Now only the ending "*aḥ*" of $S_7$ words are combinable. Hence, clearly, $S_2$ and $S_7$ words cannot combine. Hence, if $n_1 <= n_2$, then no reduction with $S_7$ words can take place. Hence $r_3$ for such a scenario is zero.

When $n_1 > n_2$, then $n_1 - n_2$ words of the $S_1$ type remain after the $r_1$ calculation. For example, we may have the word *ahaṁ* of $S_1$ remaining. Thus, *kṛṣṇaḥ + ahaṁ = kṛṣṇo'haṁ*, which is a reduction. Similarly, *kṛṣṇaḥ + idānīṁ = kṛṣṇa idānīṁ* which is again a reduction. The number of such reductions is min($n_1-n_2$, $n_7$) because there will be a maximum of as many reductions as there are pairs of the remaining $n_1-n_2$ words and $n_7$ words.

*D. Combined Formula for Reductions and **Min***

Combining the formulae for $r_1$, $r_2$ and $r_3$, we arrive at the following formula for **r**, the maximum number of possible reductions for the given set of words:

if ($n_1 = n_2$) or ($n_1 > n_2$ and $n_7 = n_1 - n_2$ and $n_5 > 0$) then

$r = n_1 + n_3 + n_5 - 1$;

else if ($n_1 < n_2$) or ($n_1 > n_2$ and [($n_7 > n_1 - n_2$) or ($n_7 = n_1 - n_2$ and $n_5 = 0$)]) then

$r = n_1 + n_3 + n_5$;

else

$r = n_2 + n_3 + n_5 + n_7$;

end if

Using **r**, we calculate **Min**, the minimum possible number of syllables as **Min = Max − r**.

V. SAMPLE SCENARIO OF PERFORMANCE ENHANCEMENT USING THE **MAX-MIN** BAND

Now consider an example where **Max** = 48. This means that in a single *pāda* or quarter of the verse, there can be a maximum of 48/4 = 12 syllables. Let us assume that by the above formulations, we arrive at **r** = 4 for this particular case. Then **Min** = 44, and hence the minimum number of syllables per *pāda* is 11. Thus, we need to only search for metres with 11 or 12 syllables in a *pāda*. This reduces the target search space of metres to match with, by more than 85%.

VI. CONCLUSIONS

The problem of automatic conversion of prose to poetry has never been tackled before in the literature. This solution is therefore the first of its kind. Further, the algorithm presented is comprehensive and yet efficient due to the reduction formula proposed through the **Max-Min** Band method. The approach also facilitates the giving of suggestions to the poet, in case the prose given cannot be set to poetry, for it finds the closest match. Lastly, and significantly, the application also easily lends itself, through available English speech recognition interfaces that come packaged with operating systems itself, for use by the visually impaired.

## Author Profile

**Dr. Rama N.** completed B.Sc. (Mathematics), Master of Computer Applications and Ph.D. (Computer Science) from the University of Madras, India. She served in faculty positions at Anna Adarsh College, Chennai and as Head of the Department of Computer Science at Bharathi Women's College, Chennai, before moving on to Presidency College, Chennai, where she currently serves as Associate Professor. She has 20 years of teaching experience including 10 years of postgraduate (PG) teaching, and has guided 15 M.Phil. students. She has been the Chairperson of the Board of Studies in Computer Science for UG, and Member, Board of Studies in Computer Science for PG and Research at the University of Madras. Current research interests: Program Security. She is the Member of the Editorial cum Advisory Board of the Oriental Journal of Computer Science and Technology.

**Meenakshi Lakshmanan** Having completed B.Sc. (Mathematics), Master of Computer Applications at the University of Madras and M.Phil. (Computer Science), she is currently pursuing Ph.D. (Computer Science) at Mother Teresa Women's University, Kodaikanal, India. She is also pursuing Level 4 Sanskrit (*Samartha*) of the *Samskṛta Bhāṣā Pracāriṇī Sabhā*, Chittoor, India. Starting off her career as an executive at SRA Systems Pvt. Ltd., she switched to academics and currently heads the Department of Computer Science, Meenakshi College for Women, Chennai, India. She is a professional member of the ACM and IEEE.